\documentclass[runningheads]{llncs}
\usepackage{graphicx}
\usepackage{algorithm}
\usepackage{algpseudocode}
\usepackage{booktabs}
\usepackage[dvipsnames]{xcolor}
\usepackage[misc]{ifsym}
\usepackage{amsfonts}

\begin{document}
\title{Multi-layer Domain Adaptation for Deep Convolutional Networks}
%
%
\author{Ozan Ciga$^{(\textrm{\Letter})}$, Jianan Chen \and
Anne Martel}
\authorrunning{O. Ciga et al.}
%
\institute{Medical Biophysics, University of Toronto, Toronto, Canada\\
\email{ozan.ciga@mail.utoronto.ca}
}

\maketitle              

\begin{abstract}
Despite their success in many computer vision tasks, convolutional networks tend to require large amounts of labeled data to achieve generalization. Furthermore, the performance is not guaranteed on a sample from an unseen domain at test time, if the network was not exposed to similar samples from that domain at training time. This hinders the adoption of these techniques in clinical setting where the imaging data is scarce, and where the intra- and inter-domain variance of the data can be substantial. We propose a domain adaptation technique that is especially suitable for deep networks to alleviate this requirement of labeled data. Our method utilizes gradient reversal layers \cite{ganin2014unsupervised} and  Squeeze-and-Excite modules \cite{hu2018squeeze} to stabilize the training in deep networks. The proposed method was applied to publicly available histopathology and chest X-ray databases and achieved superior performance to existing state-of-the-art networks with and without domain adaptation. Depending on the application, our method can improve multi-class classification accuracy by 5-20\% compared to DANN introduced in \cite{ganin2014unsupervised}. 
\end{abstract}

\section{Introduction}

Deep learning models have achieved great success in recent years on computer vision tasks. Fully convolutional networks (FCNs) consistently achieve the state-of-the-art performance in various tasks such as segmentation, classification and detection. Despite their success, however, FCNs usually require large amounts of labeled data from the domain in which the network will be deployed. As network architectures become deeper with more trainable parameters, the requirement for large amounts of data is further exacerbated as the networks are more prone to overfitting. This leads to a need for even larger amounts of data to achieve generalization. Furthermore, regardless of the size or the domain diversity of the training set, there is no performance guarantee on an unseen dataset from a domain that the network was not exposed to at training time. These issues are especially problematic in medical image analysis, as the labeled data is scarce due to the tedious and expensive data annotation process, and a large distributional shift can be observed even if data comes from the same source.  


Several methods, including network weight regularization, semi-supervised approaches \cite{baur2017semi}, meta-learning \cite{maicas2018training}, and domain adaptation \cite{ganin2014unsupervised} have been proposed to improve generalization performance on unseen datasets. In the present work, we will focus on the domain adaptation. These methods aim to leverage large amounts of cheap unlabeled data from a target domain to improve generalization performance using small amounts of labeled data. In past work, \cite{shimodaira2000improving} proposed correcting covariate shift between domains by reweighting samples from source domain to minimize the discrepancy between source and the target. This approach was later improved by minimizing distances between feature mappings of source and target domains instead of the samples itself \cite{ganin2014unsupervised}. Further modifications were proposed later that improved the benchmark performances such as tri-learning, which assumes high confidence predictions are correct \cite{saito2017asymmetric}, or leveraging the cluster assumption, in which the decision boundaries based on the modified feature representations should not cross the high density data regions \cite{shu2018dirt}. 

In the present article we propose a simple, robust method that
requires minimal modifications to an existing deep network to achieve domain adaptation. Our model repurposes Squeeze-and-Excite blocks, introduced by \cite{hu2018squeeze} for feature selection, to perform domain classification in the intermediate layers of a large network. We use the ``squeeze" operation to get a summary statistic at the end of each convolutional block, and use a domain adaptation technique \cite{ganin2014unsupervised} to extract domain-independent features at each layer. The ``excitation network" is repurposed to perform domain classification. We extend this method by matching distributions of source and target features at each layer via minimizing the Wasserstein distance. 
\section{Methods} 
Due to its conceptual simplicity, we will build our model on top of the gradient reversal layer (GRL) based domain adaptation, which was first introduced in \cite{ganin2014unsupervised}. In an FCN, convolutional layers extract salient features layer by layer as the feature maps shrink in spatial size and expand in semantic (depthwise) information. Once enough abstraction on the image is achieved, features $\mathbf{f}$ 
are flattened and typically fed into a few fully connected layers to perform the task objective, e.g., classification. As the network usually optimizes a minimization objective, extracted features may (and are likely to) overfit to the domain-specific noise. Domain adaptation via gradient reversal aims to alleviate this by attaching another classifier to the input $\mathbf{f}$, which simultaneously optimizes an adversarial objective: Given $\mathbf{f}$, it tries to minimize the domain classification loss $L_d$ between $N$ samples of the domain classifier with parameters $\theta_d$ while trying to maximize this loss with respect to the feature extractor (with parameters $\theta_f$) of the original FCN. In effect, this procedure aims to remove the learned features which are domain-specific, while forcing the network to retain the domain-independent features with error gradient signals $\frac{\partial L_y}{\partial \theta_y}$ and $\frac{\partial L_y}{\partial \theta_f}$, where $\theta_y$ are the parameters of the label classifier. 

\begin{figure}
  \centering
    \includegraphics[scale=0.54]{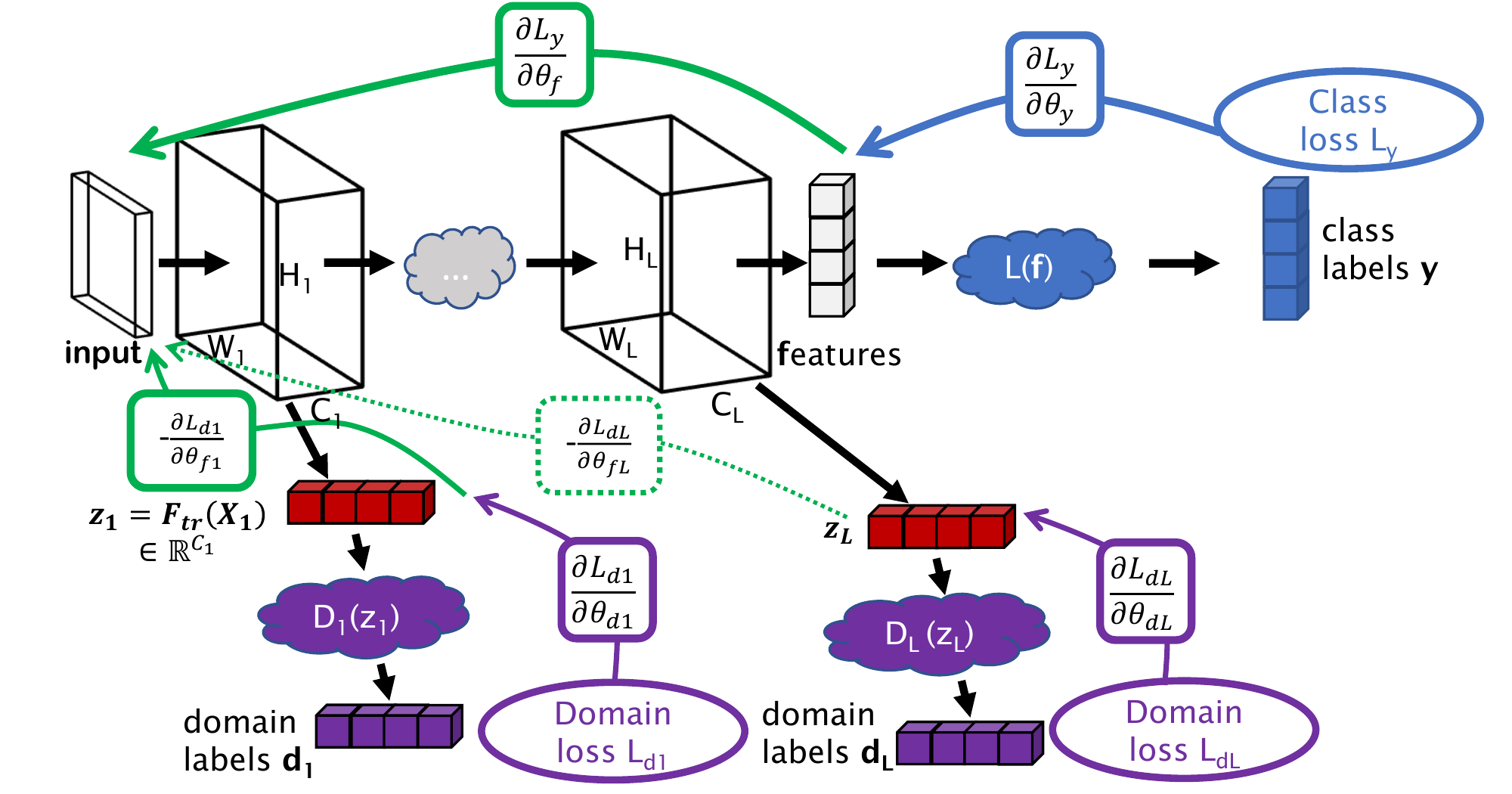}
\caption{Proposed modification to the DANN architecture.}\label{fig: lgrl_block}
\end{figure}

In \cite{ganin2014unsupervised}, domain adaptation is achieved by backpropagating the negative binomial cross-entropy loss of the domain classifier network. Features from the last layer prior to the fully connected classification layers are used as inputs to the domain classifier network. We note several problems with this approach: (1) as the network depth increases, the error signal from the domain classifier will tend to vanish, or will be insufficient to remove domain specific features in the earlier layers, (2) given feature maps $\mathbf{X}^i$ and $\mathbf{X}^j$ where $i<j$, it becomes more challenging for the network to extract domain-independent features for $\mathbf{X}^j$ if the features from $\mathbf{X}^i$ are domain dependent, (3) even if domain specific features in map $\mathbf{X}^i$ somehow are discarded in the later layers, the encoding of these features into map $\mathbf{X}^i$ results in capacity underuse of the network, (4) even with the adversarial training objective which forces the preservation of salient features, it is likely for a high capacity network to employ arbitrary transformations on the target samples to match source and target distributions (for a formal derivation, see Appendix E of \cite{shu2018dirt}). For simple tasks that do not require deep networks, vanishing gradients or accumulation of domain dependent features across layers do not affect the performance as much. However, in more complex medical imaging analysis tasks, larger networks tend to perform better; hence, the domain adaptation techniques are more likely to suffer from aforementioned issues. We aim to alleviate this by regulating extracted features at each layer simultaneously by attaching a domain classifier at the end of the layer (see Figure \ref{fig: lgrl_block}), or by performing unsupervised matching of distributions at each layer.

Given a feature map $\mathbf{X} \in \mathbb{R}^{H' \times W' \times C'}$, we transform $\mathbf{X}$ into $\mathbf{z} \in \mathbb{R}^{C'}$ by average pooling, i.e., $z_k = \frac{1}{H' \times W'} \sum_{i=1}^{H'}\sum_{j=1}^{W'} u_k(i,j)$, where $u_k(i,j)$ indexes the $(i,j)^{th}$ element of the response to the $k^{th}$ kernel of the map $\mathbf{X}$, and $z_k$ is the $k^{th}$ element of the vector $\mathbf{z}$. We will use the shorthand $\mathbf{f}_{tr}(\mathbf{X}_i)=\mathbf{z}_i$ for the transformation of map $\mathbf{X}_i$ (feature maps of layer $i$) into $\mathbf{z}_i$, which is coined as the ``squeeze" operation by \cite{hu2018squeeze}. Although $\mathbf{z}_i$ itself is not enough for downstream tasks such as classification or segmentation, it may contain enough information to differentiate between two samples at a given layer. Given this information, we aim to be able to perform domain adaptation at each layer, rather than just the final feature map representation at the end of the network.

\subsection{Gradient reversal layer based domain adaptation}

Analogous to \cite{ganin2014unsupervised}, we add domain classifiers at the end of each feature map $\mathbf{X}_i$. By interfering at the intermediate layers, we aim to extract robust features that are invariant to the training domain using the supervision signal. The network is then trained simultaneously for the domain adaptation along with the original objective. We denote this as layer-wise domain-adversarial neural network, or \textit{L-DANN}, as our model is based on DANN \cite{ganin2014unsupervised}. 

The mini domain classifier network for each layer has the same structure for each layer $C'$, but with varying number of parameters (see Table \ref{tab:discriminator_architecture}, $r$ indicates the reduction ratio). As the earlier layers in convolutional networks tend to extract more high level information such as texture patterns and edges, we increase the complexity of the domain classifier network progressively, proportional to the depth of the feature map $\mathbf{X}_i$. Given $N$ domains, the domain classifier network maximizes the $N$-class cross entropy loss via backpropagation to obscure domain information by removing the features from the map $\mathbf{X}_i$. 

\subsection{Wasserstein distance based domain adaptation}
Instead of using the domain labels directly, we can also achieve domain adaptation by interpreting $\mathbf{z}_i$ as samples drawn from different distributions. Given two domains $\mathcal{X}^s$, $\mathcal{X}^t$, with $\mathbf{z}_{s}$ and $\mathbf{z}_{t}$ are samples drawn from $\mathcal{X}^s$ and $\mathcal{X}^t$, respectively, our objective is $
    \min_{\mathbf{z}_{s}, \mathbf{z}_{t}} d(\mathbf{z}_{s}, \mathbf{z}_{t})
$ where $d(\cdot, \cdot)$ is an arbitrary distribution divergence. For our experiments, we use the Wasserstein-1 distance, also known as the Earth mover's distance, due to its stability in training \cite{arjovsky2017wasserstein}. In order to stabilize the training further, we will use the method described in \cite{gulrajani2017improved} to ensure Lipschitz constraint on the critic, as opposed to the gradient clipping method suggested in \cite{arjovsky2017wasserstein}. We use the term ``critic" as opposed to discriminator/classifier, to be consistent with \cite{arjovsky2017wasserstein,gulrajani2017improved}. The procedure is summarized in Algorithm \ref{alg:wasserstein}, we omit the details for brevity, and refer the interested reader to \cite{gulrajani2017improved}. In the upcoming sections, we will refer to this method as \textit{L-WASS}, or layer-wise Wasserstein. 

\begin{algorithm}
   \caption{Unsupervised domain adaptation via Wasserstein distance with gradient penalty for feature matching. Squeezed feature map from layer $l$ is $\mathbf{z}^k_l$, given input $x^k$. The objective loss is $\mathcal{L}_{obj}$ (e.g., cross-entropy for classification). }\label{alg:wasserstein}
    \begin{algorithmic}[1]
    \Require{source $X^s$ with samples $x^s$ and labels $y^s$, target $X^t$, number of critic iterations $n_{critic}$ per generator iteration, batch size $m$, learning rates $\alpha_{1,2}$, gradient penalty coefficient $\lambda$, initial parameters for the critic and the neural network for the objective, $\theta_d$, $\theta_f$}
    \Repeat
    \For{each layer $j$}
        \For{t=1 to $n_{critic}$}
            \For{i=1 to $m$}
            	\State Sample $(x_i^s, y_i^s) \sim X^s,\ \  x_i^t \sim X^t$, a random number $\epsilon \sim U[0, 1]$
                \State $\mathbf{z}^b_j \leftarrow \epsilon \mathbf{z}^s_j + (1-\epsilon) \mathbf{z}^{t}_{j}$
                \State $L^{(i)} \leftarrow  D_{j}(\mathbf{z}^s_j) - D_{j}(\mathbf{z}^t_j) - \lambda (||\nabla_{\mathbf{z}^b_j} D_{j}(\mathbf{z}^b_j)||_2 - 1)^2$
                \State $L_{obj}^{(i)} \leftarrow  \mathcal{L}_{obj} (x_i^s, y_i^s) - \lambda (||\nabla_{\mathbf{z}^b_j} D_{j}(\mathbf{z}^b_j)||_2 - 1)^2$
            \EndFor
            \State $\theta_d \leftarrow SGD(\nabla_d \frac{1}{m}\sum_{i=1}^m L^{(i)}, \theta_d, \alpha_1)$
    \EndFor
\EndFor
    \State $\theta_f \leftarrow SGD(\nabla_f \frac{1}{m}\sum_{i=1}^m L_{obj}^{(i)}, \theta_f, \alpha_2)$
    \Until{$\theta_f$ converges}
\end{algorithmic}
\end{algorithm}

\section{Experimental results}

\subsection{Implementation details}

We do not use any padding or bias in the convolutional layers described in Table \ref{tab:discriminator_architecture}, and use the reduction ratio $r=16$ for all the layers. We use ResNet architecture enhanced with Squeeze-and-Excite blocks as our task objective network with varying number of layers depending on the task. Contrary to \cite{ganin2014unsupervised}, we do not use a constant $\lambda$ to scale $\frac{\partial L_d}{\partial \theta}$, nor do we use annealing to stabilize the training. We use stochastic gradient descent (SGD) optimizer in all domain classifier, critic, and the objective network with the learning rate 0.001, momentum 0.9 and weight decay of 0.0001. We have tried updating the domain classifier and critic parameters with and without freezing the preceding layers and observed simultaneous training achieves superior performance. We perform 10 runs per experiment, and report the mean accuracy $\pm$ the standard deviation. All experiments are run for 100 epochs regardless of the network architecture or the data, and we use the model with the highest validation accuracy achieved in the last 30 epochs for testing, to avoid selecting a model that achieved high accuracy randomly, and has actually converged.

\begin{table}
\centering
    \begin{tabular}{c c c c c}
     \ & Input shape & Kernel size & Output shape \\
    \midrule
      $\mathbf{f}_{tr}(\mathbf{X}_i)$ & [1 $\times$ 1] $\times$ C'\ \ \ \ & - & - \\
      Conv & [1 $\times$ 1] $\times$ C'\ \ \ \  & [1 $\times$ 1] $\times$ C'/r & [1 $\times$ 1] $\times$ C'/r  \\
      ReLU & [1 $\times$ 1] $\times$ C'/r & - & [1 $\times$ 1] $\times$ C'/r \\
      Conv & [1 $\times$ 1] $\times$ C'/r & [1 $\times$ 1] $\times$ $N'$ & $N'$ \\
    \bottomrule
    \end{tabular}
    \caption{Domain classifier/critic \textit{D($\mathbf{f}_{tr}(\mathbf{X}_i)$)}. The final output shape $N'$ depends on the architecture used: For L-DANN, we use $N'=N$, or number of classes, and for L-WASS, we use $N'=C'$, number of input channels to perform distribution matching.}
    \label{tab:discriminator_architecture}
\end{table}
 
\subsection{Effect of layer-wise domain adaptation on small networks} 

In order to determine whether layer-wise domain adaptation improves results on networks with a small number of layers, we use the MNIST handwritten digits, MNIST-M (MNIST blended with random RGB color patches from the BSDS500 dataset), and the SVHN (street view house numbers) to perform digit classification given an image which contains a single digit. SVHN has more variation within the dataset; hence classifying SVHN digits is considered to be more challenging than MNIST or MNIST-M. For all experiments, we use $\sim 60000$ images per dataset for training, and $\sim 10000$ for testing. We use a single 2-layer neural network, \textit{MNIST architecture} defined in \cite{ganin2014unsupervised}, enhanced with batch normalization prior to ReLU layers. As we do not optimize the architecture depending on the dataset, or the direction of the adaptation, our results should only be interpreted within the context of Table \ref{tab:compare_small}, and not to the results reported in \cite{ganin2014unsupervised}. As the MNIST architecture is not convolutional, we use the domain classifier given in MNIST architecture for each layer. For L-WASS, the classifier remains the same, with the exception that the number of output elements are 100, to achieve more meaningful matching of distributions. Although the performance of L-DANN remains comparable to DANN, L-WASS fails to converge for the simplest experiment, hinting that for simple distributions, layer-wise Wasserstein distribution matching is not suitable.

\begin{table}
\centering
    \begin{tabular}{c c c c c}
      Method & MNIST$\rightarrow$MNIST-M & MNIST$\rightarrow$SVHN & SVHN$\rightarrow$MNIST \\
    \midrule
      No adaptation & 58 $\pm$2 & 27.9$\pm$5.41 & 77$\pm$0.96 \\
      DANN & 90.8 $\pm$1.06 & 27.7$\pm$1.43 & 46.1$\pm$2.27 \\
      L-DANN & 90.5 $\pm$0.12 & 22.8 $\pm$1.72 & 53.8 $\pm$2.22 \\
      L-WASS & N/C & 21.0$\pm$2.11 & 71.2$\pm$0.91 \\
    \bottomrule
    \end{tabular}
        \caption{Comparison between DANN, L-DANN and L-WASS for smaller networks. N/C: Network did not converge.}\label{tab:compare_small}
\end{table}

\subsection{Effect of model complexity on domain adaptation}

We test our method on another modality, namely on chest X-ray images acquired from two separate institutions in USA, and in China that are classified into normal patients as well as patients with manifestations of tuberculosis \cite{openi}. The datasets vary in resolution, quality, contrast, positive to negative samples ratio, and the number of samples. In addition, each dataset has separate watermarks and descriptive texts in different parts of the X-rays, which are known to degrade performance in neural networks. The first dataset consists of 138 images, which we refer to as \textit{S}, or small, and the second dataset consists of 662 image, which we refer to as \textit{L}, or large. In order to show that our method performs better with deeper architectures, we compare two architectures: SE-ResNet-101 (49.6 million trainable parameters) and SENET 154 (116.3M). Results are shown in Table \ref{tab:compare_large}. Note that although DANN slightly outperforms L-WASS in one of the experiments, its performance is not consistent. In some settings, it performs worse than networks without any domain adaptation, and even fails to converge for the deepest setting. In contrast, both L-DANN and L-WASS consistently perform better than the no domain adaptation baseline. The utility of using a deeper architecture can be observed in the $S\rightarrow$L direction, where we gain up to $\sim 7\%$ in accuracy, for $L-DANN$ setting. In other words, deeper networks can help better generalize to larger datasets given a small labeled dataset, which is often the case in the clinical setting.

\begin{table}
\centering
    \begin{tabular}{ c| c| c| c c c c}
    Architecture & Source$\rightarrow$ Target & Method & Precision & Recall & F1-score & Accuracy \\
    \midrule
      SE-ResNet-101 & L$\rightarrow$S & No adaptation & 100. & 18.9 & 31.8 & 65.9 \\
       &  & \textit{DANN} & 80.9 & 65.5 & 72.4 & 79. \\
       &  & \textit{L-DANN} & 88.1 & 63.8 & 74. & 81.2 \\
       &  & \textit{L-WASS} & 91.1 & 53.4 & 67.3 & 78.3 \\
       \midrule
       & S$\rightarrow$L & No adaptation & 68.7 & 72.6 & 70.6 & 69.3 \\
       &  & \textit{DANN} & 71.8 & 67.6 & 69.6 & 70.1 \\
       &  & \textit{L-DANN} & 72.6 & 73.7 & 73.1 & 73.9 \\
       &  & \textit{L-WASS} & 70.9 & 76.1 & 73.4 & 72.1 \\
    \midrule
      SENET 154 & L$\rightarrow$S & No adaptation & 100. & 3.4 & 6.6 & 59.4 \\
       &  & \textit{DANN} & 90.9 & 51.7 & 65.9 & 77.5 \\
       &  & \textit{L-DANN} & 100. & 43.1 & 60.3 & 76.1 \\
       &  & \textit{L-WASS} & 90.9 & 68.9 & 78.4 & 84.1 \\
       \midrule
       & S$\rightarrow$L & No adaptation & 79.3 & 65.1 & 71.5 & 73.7 \\
       &  & \textit{DANN} & N/C & N/C & N/C & N/C \\
       &  & \textit{L-DANN} & 75.1 & 84.5 & 79.5 & 80.9 \\
       &  & \textit{L-WASS} & 88.8 & 75.1 & 81.6 & 81.3 \\
      \bottomrule
    \end{tabular}
   \caption{Comparison between DANN, L-DANN and L-WASS for deeper networks. }\label{tab:compare_large}
\end{table}

\subsection{Domain adaptation for feature regularization}

We also test our method on the BACH (BreAst Cancer Histopathology) challenge \cite{bach2018}. This challenge is composed of classification of patches extracted from whole-slide images (WSI) into 4 classes (normal, benign, in-situ, and invasive cancer) and segmentation of the WSI into these classes. As it is not uncommon to achieve $\sim 90\%$ accuracy on the classification part, we turn our attention to the segmentation. There are 10 labeled + 20 unlabeled WSI for training, and 10 for testing. Given the stain variation among WSI, we are using the unlabeled 20 images for stain normalization, and for source (i.e., the institution, scanner or the hospital) agnostic feature extraction. In this respect, the domain adaptation acts as a regularizer on extracted features, retaining only the features which are common in both domains. We train the same network, SE-ResNet-50, without domain adaptation, with \textit{L-DANN} module, with \textit{L-WASS}, and with \textit{DANN}, and achieve scores (as defined in \cite{aresta2019bach}, which penalizes false negatives, or incorrect ``normal" class, more than false positives, or any of the remaining three classes) \textbf{0.63}, \textbf{0.68}, \textbf{0.66}, \textbf{0.65}, respectively. Note that the $2^{nd}$ best score on the public leaderboard is \textbf{0.63}. 

\section{Conclusions}

We presented a novel domain adaptation method for fully convolutional networks that can alleviate the requirements for large amounts of data, especially in deep networks. Our method is simple, requires minimal amount of modification to the original network architecture, adds small overhead to the training cost, and is cost-free in test time. We tested our method with multiple public medical imaging datasets and showed promising gains on multiple baseline networks. 

\section{Acknowledgments}
This work was funded by Canadian Cancer Society (grant $\#$ 705772) and NSERC RGPIN-2016-06283.

%
%
\bibliographystyle{splncs04}
\bibliography{mybibliography}

\end{document}